\documentclass[11pt]{article}

\usepackage[final]{acl}

\usepackage{times}
\usepackage{latexsym}

\usepackage[T1]{fontenc}

\usepackage[utf8]{inputenc}

\usepackage{microtype}

\usepackage{inconsolata}

\usepackage{graphicx}
\usepackage{booktabs}
\usepackage{enumitem}
\usepackage{float}
\usepackage{booktabs}
\usepackage[table]{xcolor}

\usepackage{caption}
\usepackage{multirow}
\usepackage{subcaption}
\usepackage{enumitem}
\usepackage{placeins}
\usepackage{longtable}
\usepackage{fancyvrb}
\usepackage[most]{tcolorbox}
\tcbuselibrary{breakable}
\usepackage{tikz}
\usetikzlibrary{arrows.meta, positioning, shapes, fit}
\usepackage{xcolor}
\definecolor{purple}{RGB}{128,0,128}
\definecolor{teal}{RGB}{0,128,128}
\newtcolorbox{promptblock}{
  colback=gray!4,
  colframe=gray!35,
  boxrule=0.35pt,
  arc=0pt,
  left=5pt,
  right=5pt,
  top=4pt,
  bottom=4pt,
  breakable,
  enhanced
}

\title{HypoKG: Evidence-Disciplined Biomedical Hypothesis Generation Beyond Endpoint Knowledge}

\author{
Dominic Okonkwo,
Adetayo Okunoye,
Ismailcem Budak Arpinar \\
School of Computing \\
University of Georgia \\
\texttt{\{dominic.okonkwo, adetayo.okunoye, budak\}@uga.edu}
}

\begin{document}
\maketitle
\begin{abstract}
Large language models (LLMs) can generate biomedical hypotheses, but it remains unclear whether they truly reason from scientific evidence or simply produce convincing-sounding ideas. To study this, we combine three major biological databases: the Kyoto Encyclopedia of Genes and Genomes (KEGG), Rhea, and UniProt, into a unified biochemical knowledge graph and construct a benchmark of 550 paths connecting enzyme sources to rare disease endpoints, yielding 13,200 hypotheses from six LLMs under four conditions varying the biological information each model receives: source enzyme only, full biological path, or source and disease endpoint only. Hypotheses are scored using an expert-derived five-criterion rubric on a 1--5 scale per criterion. We find that models given both the source and disease endpoint often produce the highest-scoring hypotheses, showing that LLMs can generate compelling ideas from minimal information. However, these hypotheses are less grounded in the evidence. In contrast, models given the full biological path generate hypotheses more consistent with known mechanistic relationships. We call this \textit{evidence-disciplined reasoning}. To confirm this effect, we shuffled intermediate path steps while keeping endpoints fixed. Evidence grounding dropped significantly ($\Delta = -0.793$, $p < 0.001$), confirming models genuinely used path structure during reasoning. Our findings show that knowledge graphs support hypothesis generation in two ways: they identify biological endpoint pairs absent from the literature, and their mechanistic paths guide how LLMs reason between them.
\end{abstract}

\section{Introduction}

Biomedical hypothesis generation requires more than plausible associations: models must produce mechanistic claims proportionate to the evidence available. Large language models (LLMs) have shown strong scientific reasoning abilities \citep{Brown2020,Chowdhery2023} and are increasingly used to generate
hypotheses \citep{Yang2024,Si2024,Liu2025,AbdelRehim2025}, yet a key question remains unresolved: when LLMs are given knowledge graph (KG) evidence connecting a biological source entity to a disease endpoint, does that structure change the hypothesis produced, or do models exploit parametric memory and generate plausible endpoint-driven explanations?

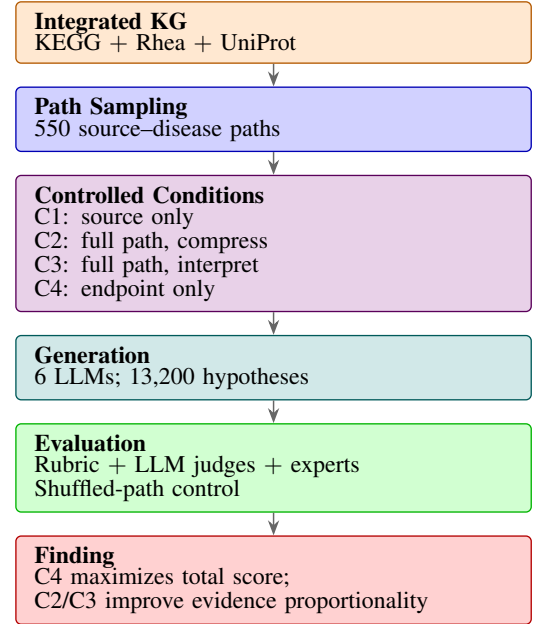
\begin{figure}[t]
\centering
\begin{tikzpicture}[
  font=\footnotesize,
  box/.style={
    rounded corners=2pt,
    draw=#1!70!black,
    fill=#1!18,
    line width=0.5pt,
    inner xsep=7pt,
    inner ysep=4pt,
    align=left,
    text width=0.82\columnwidth
  },
  arrow/.style={-{Stealth[length=2.0mm]}, line width=0.55pt, draw=black!60},
  node distance=3mm
]

\node[box=orange] (kg) {%
  \textbf{Integrated KG}\\[-2pt]
  KEGG $+$ Rhea $+$ UniProt
};

\node[box=blue, below=of kg] (paths) {%
  \textbf{Path Sampling}\\[-2pt]
  550 source--disease paths
};

\node[box=purple, below=of paths] (conds) {%
  \textbf{Controlled Conditions}\\[-2pt]
  C1: source only\\[-1pt]
  C2: full path, compress\\[-1pt]
  C3: full path, interpret\\[-1pt]
  C4: endpoint only
};

\node[box=teal, below=of conds] (gen) {%
  \textbf{Generation}\\[-2pt]
  6 LLMs; 13,200 hypotheses
};

\node[box=green, below=of gen] (eval) {%
  \textbf{Evaluation}\\[-2pt]
  Rubric $+$ LLM judges $+$ experts\\[-1pt]
  Shuffled-path control
};

\node[box=red, below=of eval] (finding) {%
  \textbf{Finding}\\[-2pt]
  C4 maximizes total score;\\[-1pt]
  C2/C3 improve evidence proportionality
};

\draw[arrow] (kg) -- (paths);
\draw[arrow] (paths) -- (conds);
\draw[arrow] (conds) -- (gen);
\draw[arrow] (gen) -- (eval);
\draw[arrow] (eval) -- (finding);

\end{tikzpicture}
\vspace{-2pt}
\caption{Overview of HypoKG: an integrated KEGG--Rhea--UniProt benchmark for testing whether full KG paths improve evidence-grounded biomedical hypothesis generation beyond endpoint knowledge alone.}
\vspace{-0.8em}
\vspace{-8pt}
\end{figure}

This question is especially important in biomedicine, where plausible but weakly supported mechanisms can mislead experimental design. Resources such as KEGG \citep{Kanehisa2000}, Rhea \citep{Bansal2022}, and UniProt \citep{UniProt2023} expose multi-hop cross-domain paths connecting enzyme-kinetics source entities to rare disease endpoints, connections often absent from published literature. However, the presence of a KG path does not guarantee that an LLM uses it mechanistically. A model may instead ignore intermediate structure and construct a fluent explanation from the source and disease names alone.

Existing KG-grounded biomedical generation work has focused on factual question answering \citep{Soman2024}, hypothesis generation with hallucination reduction \citep{Xiong2024}, or truthfulness verification \citep{Wen2024}, where evaluation relies on known answers. Hypothesis generation is different: quality is multi-dimensional and KG grounding must be measured against a strong alternative, endpoint-only generation. Prior benchmarks \citep{Xiong2025,Kumbhar2025} do not isolate whether full mechanistic paths add value beyond revealing the source--disease endpoint pair.

We introduce \textsc{HypoKG}, a benchmark for KG-grounded biochemical hypothesis generation built from an integrated KEGG--Rhea--UniProt graph (17{,}449 nodes; 31{,}709 edges). We sample 550 multi-hop paths connecting enzyme-kinetics sources to rare disease endpoints (94.5\% absent from PubMed co-citation). Six LLMs generate 13{,}200 hypotheses across four conditions: source-only (C1), full-path compression (C2), full-path interpretation (C3), and endpoint-only (C4), the last testing whether models can generate plausible hypotheses from KG-discovered endpoints without the mechanistic path.

We evaluate hypotheses using an expert-derived five-criterion rubric with a cross-judge LLM-as-judge framework (Pearson $r=0.776$) validated by three PhD-level domain experts. The results reveal a central tension: endpoint-only prompting achieves the highest total rubric score, but full KG paths uniquely improve evidence proportionality, whether mechanistic claims are supported by path evidence rather than speculation. We call this \textit{evidence-disciplined reasoning}. A shuffled-path control confirms the effect depends on genuine use of intermediate path structure, strongest in larger and domain-specialized models. Our contributions are:
\begin{enumerate}[noitemsep, itemsep=0pt, topsep=4pt, partopsep=0pt, parsep=0pt, leftmargin=*]
    \item \textbf{HypoKG benchmark.} A 550-path biochemical hypothesis generation benchmark with 13,200 hypotheses across six LLMs and four controlled grounding conditions, including the first endpoint-only ablation for KG-grounded hypothesis generation.
    \item \textbf{Evidence-disciplined reasoning.} Full KG paths do not maximize aggregate judged quality; they specifically improve evidence proportionality, constraining models toward grounded mechanistic claims. This finding is validated by expert human evaluation across 55 paths.
    \item \textbf{Path structure utilization control.} A shuffled-path intervention showing that evidence proportionality gains depend on genuine intermediate path structure, with significant drops when intermediate nodes are permuted while endpoints are held fixed (C2: $\Delta = -0.793$, $p < 0.001$).
\end{enumerate}

\section{Related Work}

\textbf{Scientific Hypothesis Generation with LLMs.} Recent work has explored LLMs for hypothesis generation across open-domain discovery, data-driven induction, and multi-agent biomedical settings \citep{Yang2024,Zhou2024,Xu2025}, with human studies showing LLM-generated ideas are judged more novel than expert-written ones though with feasibility trade-offs \citep{Si2024}. Other approaches combine causal KGs or literature with empirical data, showing that integrating theory-driven and data-driven signals outperforms either alone \citep{Tong2024,Borrego2024,Liu2025}. Unlike these studies, HypoKG explicitly controls what grounding information is available to the model, isolating full KG paths from endpoint knowledge alone.

\paragraph{KG-Grounded Biomedical Reasoning.} Large-scale biomedical KGs including Hetionet, PrimeKG, SPOKE, and RTX-KG2 have demonstrated the value of knowledge integration for translational reasoning and precision medicine \citep{Himmelstein2017,Chandak2023,Morris2023,Wood2022}. KG-grounded LLM approaches such as KG-RAG have shown benefits for biomedical question answering and explainable generation, while graph-path methods support relation prediction \citep{Soman2024,Shi2025,Bakal2018}. However, most work treats KG access as retrieval support rather than testing how different forms of KG evidence affect open-ended hypothesis generation. HypoKG evaluates multi-hop biochemical paths from an integrated KEGG--Rhea--UniProt graph, focusing on whether intermediate path structure changes the mechanistic claims models produce.

\paragraph{Evaluation and Benchmarking.} LLM-as-judge evaluation has become common for open-ended outputs, with \citet{Zheng2023} showing strong agreement between GPT-4 judgments and human preferences while documenting judge biases. Biomedical hypothesis benchmarks including TruthHypo \citep{Xiong2025}, BioVerge \citep{Yang2025}, and \citet{Kumbhar2025} evaluate truthfulness, agent self-evaluation, and materials discovery respectively. HypoKG differs by testing a distinct question: whether full mechanistic KG paths provide value beyond revealing the same source--disease endpoints. This endpoint-only ablation is central to separating genuine path utilization from plausible endpoint-driven generation.

\section{HypoKG Benchmark}

\subsection{Benchmark Overview}

HypoKG evaluates whether structured biochemical knowledge graph (KG) evidence changes the character of LLM-generated biomedical hypotheses. The benchmark is built around 550 multi-hop cross-domain KG paths connecting enzyme-kinetics source entities to rare disease endpoints. For each path, six LLMs generate hypotheses under four controlled prompting conditions, producing 13,200 hypotheses in total. Table~\ref{tab:hypokg_summary} summarizes the graph composition, benchmark scale, and generation setup.

Unlike benchmarks that only compare grounded generation against ungrounded prompting \citep{Xiong2025,Kumbhar2025}, HypoKG separates two roles of the KG: endpoint discovery and mechanistic path grounding. The endpoint-only condition tests whether models can generate plausible hypotheses once the source-disease pair is revealed, while the full-path conditions test whether intermediate KG structure disciplines the mechanism proposed.

\subsection{Integrated Biochemical Knowledge Graph}

We construct an integrated human biochemical KG from KEGG \citep{Kanehisa2000,Kanehisa2023}, Rhea \citep{Bansal2022}, and UniProt \citep{UniProt2023}. KEGG provides pathway topology and gene--metabolite--pathway relationships. Rhea contributes reaction-level precision including directionality and substrate--product relationships. UniProt provides curated protein annotations and enzyme--disease associations from reviewed Swiss-Prot entries.

All data are restricted to human biochemistry. KEGG pathways are filtered to organism code \texttt{hsa} and UniProt proteins to taxonomy identifier \texttt{9606}, avoiding contamination from model-organism pathways sharing enzyme nomenclature with human homologs. The resulting graph contains 17,449 nodes and 31,709 edges spanning four domain types connected by six relationship types: pathway links, reaction interactions, enzyme-commission relationships, disease associations, enzyme associations, and metabolic pathway membership.

\begin{table}[t]
\centering
\small
\setlength{\tabcolsep}{4pt}
\renewcommand{\arraystretch}{1.08}
\begin{tabular}{p{0.38\columnwidth}p{0.54\columnwidth}}
\toprule
\textbf{Component} & \textbf{Details} \\
\midrule
Integrated KG & KEGG, Rhea, UniProt \\
KG size & 17{,}449 nodes; 31{,}709 edges \\
Domain types & Enzyme kinetics; metabolic pathways; disease mechanisms; drug targets \\
Benchmark paths & 550 \\
Source entities & 417 unique enzyme-related entities \\
Terminal diseases & 366 unique disease endpoints \\
Crossing counts & 3--9 \\
Models & 6 \\
Conditions & 4 \\
Total hypotheses & 13{,}200 \\
\bottomrule
\end{tabular}
\caption{Summary of the HypoKG benchmark, including graph composition, benchmark scale, and generation setup.}
\vspace{-0.8em}
\label{tab:hypokg_summary}
\end{table}

\begin{table*}[t]
\centering
\small
\setlength{\tabcolsep}{6pt}
\renewcommand{\arraystretch}{1.12}

\begin{tabular}{
p{0.18\textwidth}
>{\raggedright\arraybackslash}p{0.25\textwidth}
p{0.48\textwidth}
}
\toprule
\textbf{Condition} & \textbf{Information provided} & \textbf{Purpose} \\
\midrule
C1: Source-only 
& Source entity only 
& Measures parametric generation without endpoint or path information. \\

C2: Full-path compression 
& Full KG path 
& Tests whether models can compress noisy path evidence into a focused mechanistic hypothesis. \\

C3: Full-path interpretation 
& Full KG path 
& Tests whether explicit mechanistic interpretation before hypothesis generation improves path-grounded reasoning. \\

C4: Endpoint-only 
& Source and terminal disease only 
& Tests whether endpoint knowledge alone explains apparent grounding gains. \\
\bottomrule
\end{tabular}

\caption{Controlled generation conditions used to isolate the effect of KG path information on hypothesis generation.}
\vspace{-0.8em}
\label{tab:conditions}
\end{table*}

\subsection{Cross-Domain Path Construction}
A cross-domain path connects an enzyme-kinetics source entity to a terminal 
disease endpoint through intermediate biochemical entities. A crossing is a 
traversal step between distinct domain types. We sample 8,000 candidate paths 
using constrained shortest-path traversal implemented with NetworkX, requiring 
3--12 nodes, 1--9 crossings, and terminal nodes restricted to named disease 
entities. Higher-crossing paths are oversampled to ensure representation of 
complex reasoning cases.

From these candidates, we select 550 benchmark paths using a greedy stratified 
filter that maximizes source-entity diversity, terminal disease coverage, and 
crossing-count coverage simultaneously. When multiple candidate paths were 
equivalent under the filter objective, tie-breaking was by \texttt{hypokg\_id} 
lexicographic order. The final benchmark contains 417 unique source entities 
and 366 unique terminal diseases, with crossing counts from 3 to 9 and path 
lengths from 4 to 11 hops.

To characterize path validity, three domain experts in biochemistry and metabolic 
disease reviewed all 550 paths. No path was judged invalid. Approximately 40\% 
were classified as directly supported, with each step reflecting a clear, 
biologically plausible relationship. The remaining paths were classified as 
requiring interpretive review: endpoints and database edges were valid, but one 
or more intermediate steps passed through high-degree annotation hubs, most 
prominently pyruvate carboxylase deficiency, aceruloplasminemia, and congenital 
lactase deficiency. These paths are retained by design. HypoKG integrates three 
databases specifically to surface cross-domain connections that no single database 
exposes alone. The intermediate ambiguity they carry is not a flaw, but a feature. 
It reflects genuine uncertainty at the frontier of biochemical knowledge and tests 
whether LLMs can reason carefully under incomplete evidence.

\subsection{Literature Prevalence Annotation}

We conduct a two-pass PubMed co-citation analysis to estimate prior literature coverage of each source--disease connection. The first pass uses exact phrase matching of the raw source symbol and terminal disease name. The second pass expands both entities with validated aliases, reviewed to remove broad or ambiguous terms. Each path is assigned to a prevalence stratum: Novel, Sparse, Emerging, Established, or Well-known. The final distribution comprises 520 Novel (94.5\%), 19 Sparse (3.5\%), 5 Emerging (0.9\%), 4 Well-known (0.7\%), and 2 Established (0.4\%) paths, confirming that HypoKG primarily evaluates hypothesis generation in the low-evidence regime where models cannot rely on well-studied source--disease associations.

\subsection{Generation Conditions}

For each KG path, models generate a testable mechanistic hypothesis connecting the source entity to a disease mechanism. Four prompting conditions systematically vary the grounding information available. 
C1 establishes a baseline and exposes disease-drift behavior, models generating hypotheses about familiar diseases unrelated to the assigned endpoint. C2 instructs the model to identify the most biologically meaningful transition rather than narrating the full path. C3 asks the model to reason about meaningful versus weak transitions before committing to a hypothesis. C4 provides only the source and terminal disease, removing all intermediate structure. Table~\ref{tab:conditions} summarizes the purpose of each condition.

C4 is the critical ablation. The source--disease pair in C4 was itself discovered through KG construction, so C4 tests whether the full mechanistic path adds value beyond endpoint discovery, not whether the KG is dispensable. All prompts require specific biological entities, a falsifiable experimental prediction, and grounding in established biochemistry. Full prompts are in Appendix~A.

\subsection{Model Suite and Generation Protocol}

We evaluate six LLMs spanning general-purpose, biomedical-specialized, 
proprietary, and open-weight systems: Claude Sonnet 4.6 (Anthropic), 
GPT-4o (OpenAI), Llama-3.3-70B \citep{Meta2024}, Qwen3-235B 
\citep{Qwen2025}, BioMistral-7B \citep{Labrak2024}, and MedGemma-27B 
\citep{Andrew2025}. This suite compares KG grounding behavior across 
scale, access type, and domain specialization. Each model generates one 
hypothesis per path per condition, producing 13,200 hypotheses total. 
All generation runs used greedy decoding where supported (temperature=0), 
with no few-shot examples and a 512-token output limit. No sampling was 
applied. API models (Claude Sonnet, GPT-4o, Llama-3.3-70B, Qwen3-235B) 
were accessed via commercial APIs. HuggingFace models (BioMistral-7B, 
MedGemma-27B) were run on A100 80GB with 8-bit and 4-bit quantization 
respectively.

Truncation at the 512-token limit was rare across models. BioMistral-7B 
had the highest rate at 3.9\% (64 of 1,650 responses), consistent with 
its tendency to repeat text before completing a hypothesis. MedGemma-27B 
had a 0.2\% rate (4 of 1,650). All other models had zero truncated 
responses. Truncated responses were included as-is and scored by the 
judge on whatever text was generated; given the low rates, truncation is 
unlikely to have meaningfully affected results.

\section{Evaluation Framework}

\subsection{Automated KG Consistency Metrics}

We compute three automated grounding metrics for all 13,200 hypotheses. Source grounding measures whether the hypothesis mentions the source entity by symbol or alias. Terminal grounding measures whether it mentions the terminal disease by name or validated alias. Path entity coverage measures the fraction of intermediate KG entities mentioned.

These metrics measure entity use, not scientific quality. A hypothesis can mention many path entities while remaining mechanistically weak, or mention few while making a coherent focused claim. We treat them as complementary grounding diagnostics rather than primary quality measures. Results are reported in Appendix~C (Table~C3).

\subsection{Expert-Derived Rubric}
We evaluate hypothesis quality using an expert-derived five-criterion rubric 
developed from structured sessions with eight PhD-level domain experts in 
biochemistry and metabolic disease. Experts scored hypotheses from a held-out 
calibration path and provided written justifications. Analysis of their responses 
yielded five criteria: path/task relevance, mechanistic specificity and 
consistency, experimental testability, non-trivial novelty, and evidence 
proportionality. Each criterion is scored 1--5 with behavioral anchors and 
worked examples, yielding total scores from 5 to 25. Criteria are grounded in 
established hypothesis quality literature, experimental testability in 
\citet{Popper1959} and \citet{Banerjee2009}, mechanistic specificity in 
\citet{Gasparyan2019} and \citet{Misra2021}, with evidence proportionality 
novel to KG evaluation and derived directly from expert consensus signals. 
Notably, the rubric was derived from expert sessions conducted before the 
endpoint-only condition (C4) was introduced, ensuring that evaluation criteria 
were not designed to favor or penalize endpoint-only generation. Full rubric 
details are in Appendix~B.

Evidence proportionality is central to HypoKG. It measures whether the 
strength of the mechanistic claim is justified by the evidence available to 
the generator, distinguishing fluent endpoint-driven speculation from 
hypotheses genuinely constrained by KG path evidence. Among the five criteria, 
evidence proportionality is the only one that directly measures whether a model 
used the provided KG path. The remaining four criteria, relevance, mechanistic 
specificity, testability, and novelty, reflect scientific quality that any 
good hypothesis should have regardless of whether a path was shown. Evidence 
proportionality is therefore the natural focus for a study asking whether KG 
structure changes how models reason.

\subsection{Cross-Judge LLM Evaluation}
All hypotheses are scored by two LLM judges using the expert-derived rubric, 
with no model evaluating its own outputs. GPT-4o serves as primary judge for 
all models except GPT-4o, whose outputs are judged by Qwen3-235B. Gemini 2.5 
Flash serves as secondary judge for all hypotheses. Claude Sonnet was excluded 
as a judge after an initial evaluation showed a 14\% content filter refusal 
rate on biochemical disease hypotheses. Judge prompts include the generation 
condition context, which is necessary for scoring evidence proportionality 
correctly; a model should not be penalized for omitting path entities that 
were never provided.

Across 13,200 hypotheses, primary and secondary judges achieve Pearson 
$r = 0.776$ agreement on total score with a mean absolute difference of 2.10 
points on the 25-point scale. Agreement is highest for experimental testability 
and lowest for evidence proportionality, reflecting the greater subjectivity 
of claim-to-evidence calibration.

\subsection{Human Expert Validation}

Three PhD-level domain experts score a stratified sample of 55 paths, 10\% of the benchmark, on evidence proportionality in a blinded within-path design. For each path, experts score all four condition outputs in randomized order with condition labels withheld. Because all four hypotheses share the same source and terminal disease, the comparison controls for path difficulty and disease familiarity. Expert scores are compared with LLM-judge evidence proportionality scores using Kendall's $W$ and Spearman's $\rho$, and we report the percentage of paths for which experts confirm the main condition ordering.

\subsection{Shuffled-Path Control}
To test whether evidence proportionality gains reflect genuine intermediate 
path use rather than endpoint priors, we introduce a shuffled-path control. 
Source and terminal endpoints are held fixed while intermediate nodes are 
permuted, disrupting mechanistic order while preserving entity identity. 
Paths with fewer than three intermediate nodes cannot be meaningfully 
permuted and are excluded, yielding 518 shuffled paths.

Models generate hypotheses from shuffled paths under full-path conditions. 
If models use ordered KG structure, shuffling should reduce evidence 
proportionality while leaving endpoint-driven plausibility and novelty 
unaffected, indicating that models respond to ordered intermediate structure 
rather than merely to the presence of biomedical entities. Because the 
full-path prompts (C2 and C3) include reasoning instructions not present 
in C4, this control also isolates the contribution of path content from 
prompt framing: the prompts are held fixed while only the path content 
changes.

\begin{table*}[t]
\centering
\footnotesize
\setlength{\tabcolsep}{8.8pt}
\renewcommand{\arraystretch}{1.03}
\begin{tabular}{llrrrrrrcc}
\toprule
\textbf{Model} & \textbf{Cond.}
& \multicolumn{5}{c}{\textbf{Rubric Scores}}
& \textbf{Total}
& \textbf{C3--C1}
& \textbf{C2/3--C4} \\
\cmidrule(lr){3-7}
& & \textbf{Rel} & \textbf{Mech} & \textbf{Test} & \textbf{Nov} & \textbf{Prop}
& & & \\
\midrule

Claude Sonnet & C1 & 1.02 & 3.24 & \textbf{4.31} & 2.89 & 1.87 & 13.32 &  &  \\
 & C2 & 3.05 & \textbf{3.73} & 3.98 & 3.04 & \textbf{3.10} & 16.89 &  &  \\
 & C3 & 2.93 & 3.68 & 4.00 & 2.89 & 2.96 & 16.46 &  &  \\
 & C4 & \textbf{3.75} & 3.70 & 4.28 & \textbf{3.98} & 2.99 & \textbf{18.70} & +3.14 & $-2.02$ \\
\midrule

GPT-4o & C1 & 1.01 & \textbf{3.86} & \textbf{4.36} & 2.69 & 1.01 & 12.93 &  &  \\
 & C2 & 3.27 & 3.51 & 4.00 & 2.76 & 2.75 & 16.29 &  &  \\
 & C3 & 3.39 & 3.71 & 3.99 & 2.94 & \textbf{2.99} & 17.01 &  &  \\
 & C4 & \textbf{4.02} & 3.53 & 4.17 & \textbf{3.65} & 2.00 & \textbf{17.37} & +4.08 & $-0.72$ \\
\midrule

Llama-3.3-70B & C1 & 1.01 & 2.88 & \textbf{4.01} & 2.74 & 1.63 & 12.26 &  &  \\
 & C2 & 2.51 & 3.02 & 3.60 & 2.60 & 2.60 & 14.33 &  &  \\
 & C3 & 2.11 & 2.85 & 3.35 & 2.31 & 2.27 & 12.88 &  &  \\
 & C4 & \textbf{3.32} & \textbf{3.03} & 3.89 & \textbf{3.55} & \textbf{2.81} & \textbf{16.59} & +0.62 & $-2.98$ \\
\midrule

Qwen3-235B & C1 & 1.02 & 3.42 & 4.47 & 2.87 & 1.91 & 13.70 &  &  \\
 & C2 & 3.39 & 3.73 & 4.05 & 3.54 & \textbf{3.41} & 18.12 &  &  \\
 & C3 & 2.98 & 3.66 & 4.03 & 3.13 & 3.04 & 16.84 &  &  \\
 & C4 & \textbf{3.72} & \textbf{3.76} & \textbf{4.52} & \textbf{4.03} & 2.98 & \textbf{19.01} & +3.14 & $-1.53$ \\
\midrule

BioMistral-7B & C1 & 1.01 & 2.36 & \textbf{3.11} & 2.40 & 1.37 & 10.24 &  &  \\
 & C2 & 2.18 & 2.05 & 1.69 & 1.86 & 1.95 & 9.74 &  &  \\
 & C3 & \textbf{2.68} & 1.92 & 1.66 & 1.88 & \textbf{2.15} & 10.29 &  &  \\
 & C4 & 2.44 & \textbf{2.38} & 2.77 & \textbf{2.72} & 2.09 & \textbf{12.41} & +0.05 & $-2.39$ \\
\midrule

MedGemma-27B & C1 & 1.00 & 2.76 & \textbf{3.99} & 2.84 & 1.51 & 12.10 &  &  \\
 & C2 & 3.04 & 3.02 & 3.52 & 2.86 & 3.02 & 15.46 &  &  \\
 & C3 & \textbf{3.32} & \textbf{3.15} & 3.48 & 3.04 & \textbf{3.27} & \textbf{16.26} &  &  \\
 & C4 & 2.99 & 3.01 & 3.88 & \textbf{3.54} & 2.60 & 16.02 & +4.16 & $-0.16$ \\

\midrule
\midrule

\rowcolor{gray!10}
\textbf{All models} & C1 & 1.01 & 3.09 & \textbf{4.04} & 2.74 & 1.55 & 12.43 &  &  \\
\rowcolor{gray!10}
 & C2 & 2.91 & 3.18 & 3.47 & 2.78 & \textbf{2.81} & 15.14 &  &  \\
\rowcolor{gray!10}
 & C3 & 2.90 & 3.16 & 3.42 & 2.70 & 2.78 & 14.96 &  &  \\
\rowcolor{gray!10}
 & C4 & \textbf{3.37} & \textbf{3.23} & \underline{3.92} & \textbf{3.58} & 2.58 & \textbf{16.68} & +2.53 & $-1.63$ \\
\bottomrule
\end{tabular}
\caption{Mean rubric scores by model and condition. Rel = Path/Task Relevance, Mech = Mechanistic Specificity, Test = Experimental Testability, Nov = Non-Trivial Novelty, and Prop = Evidence Proportionality. C3--C1 reports the KG-grounding gain; C2/3--C4 reports the gap between path-grounded and endpoint-only generation. Bold values indicate the highest value within each model for each criterion.}
\label{tab:main_results}
\vspace{-0.8em}
\end{table*}

\begin{figure}[h]
\centering
\includegraphics[width=\columnwidth]{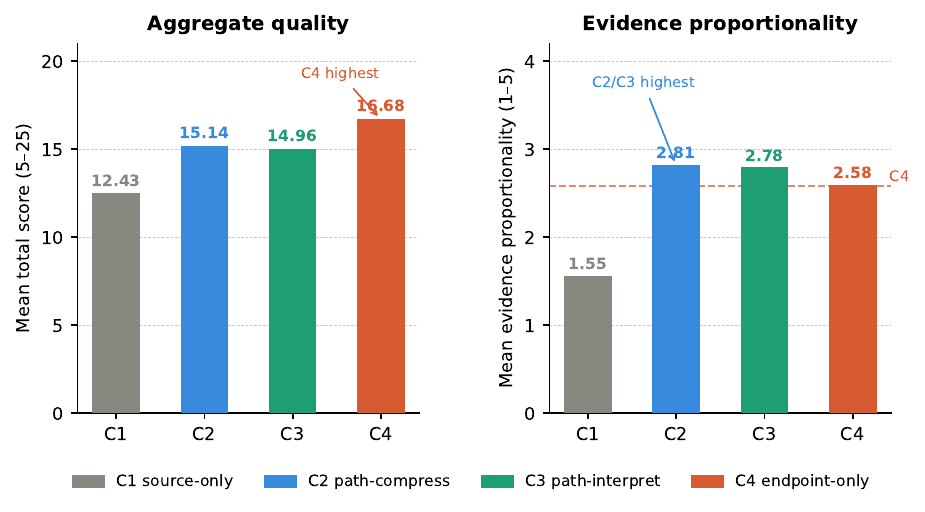}
\caption{C4 (endpoint-only) achieves the highest total rubric score, but full-path conditions (C2/C3) uniquely improve evidence proportionality - the criterion measuring whether mechanistic claims are supported by path evidence rather than speculation.}
\vspace{-0.8em}
\label{fig:main_result}
\end{figure}

\section{Results}

\subsection{Endpoint Knowledge vs. KG Grounding}
Figure~\ref{fig:main_result} provides a visual summary of the central 
pattern: endpoint-only prompting achieves the highest aggregate score, 
while full-path conditions provide their clearest advantage in evidence 
proportionality. Table~\ref{tab:main_results} reports the corresponding 
mean rubric scores by model and condition; all condition comparisons are 
evaluated over all matched path-model pairs using paired Wilcoxon 
signed-rank tests with Holm correction ($N=3{,}300$ per condition, 
aggregated across all six models), not model-specific subsets.

Endpoint-only prompting achieves the highest total rubric score across all conditions (C4: 16.68), outperforming both full-path conditions (C2: 15.14; C3: 14.96). This shows that LLMs can generate fluent, plausible, and high-scoring hypotheses from source--disease endpoint knowledge alone. However, this does not imply that KG paths are unnecessary: the endpoint pair shown in C4 was itself discovered through KG construction and path sampling, and C4's superiority on four of five criteria reflects not necessarily stronger evidence-grounded reasoning, but greater freedom to construct fluent endpoint-driven mechanisms; models shown only endpoints generate creative, relevance-maximizing hypotheses precisely because they are unconstrained by mechanistic evidence.

The key advantage of full KG paths appears in evidence proportionality. C2 and C3 significantly outperform C4 on this criterion (paired Wilcoxon, Holm corrected; C2 vs C4: $\Delta = +0.23$, $p < 0.001$, $d = 0.256$; C3 vs C4: $\Delta = +0.20$, $p < 0.001$, $d = 0.219$). Evidence proportionality is the criterion most directly measuring whether mechanistic claims are supported by the provided evidence rather than constructed speculatively from endpoint priors. The difference between C2 and C3 on evidence proportionality is not significant after correction ($p = 0.062$), consistent with their close design intent. Thus, full KG paths do not maximize aggregate judged quality; they discipline models toward claims that are more proportionate to available mechanistic evidence. We term this \textit{evidence-disciplined reasoning}.

C1 reveals a complementary failure mode we term \textit{disease-drift bias}. Source-only prompting produces near-minimum path/task relevance (1.01) while retaining high testability (4.04), indicating that models produce well-structured experimental hypotheses even when targeting the wrong disease. Without endpoint or path grounding, models default to familiar high-prevalence diseases, most commonly NAFLD, Alzheimer's disease, and type 2 diabetes, rather than the assigned rare disease endpoint. Path-grounded conditions produce mean total score gains of $+2.71$ (C2) and $+2.53$ (C3) over C1 (both $p < 0.001$, $d > 0.81$), driven primarily by large improvements in path/task relevance ($+1.90$ for C2) and evidence proportionality ($+1.26$ for C2, $d = 1.337$).

\subsection{KG Path Utilization Is Model-Dependent}

Model-level results in Table~\ref{tab:main_results} reveal a clear capability split. Stronger models, Claude Sonnet, GPT-4o, Qwen3-235B, and MedGemma-27B, show substantial gains from full-path grounding relative to source-only prompting, with C3--C1 improvements ranging from 3.1 to 4.2 points. In contrast, BioMistral-7B and Llama-3.3-70B show near-zero or weak gains (0.05 and 0.62 respectively), suggesting that smaller or less capable models cannot reliably exploit multi-hop KG structure and instead revert to parametric responses regardless of the path shown.

The C2/3--C4 gap further distinguishes models. MedGemma-27B shows the smallest gap ($-0.16$) and is the only model for which C3 outperforms C4, suggesting that biomedical domain specialization helps translate structured biochemical paths into mechanistically grounded hypotheses that match or exceed what endpoint knowledge alone provides. GPT-4o also shows a relatively small gap ($-0.72$), while Llama-3.3-70B ($-2.98$) and BioMistral-7B ($-2.39$) show large negative gaps, indicating stronger dependence on endpoint-driven generation than path-grounded reasoning.

Together, these results suggest that KG grounding is not automatically useful for all LLMs. Productive use of multi-hop biochemical paths requires sufficient model capacity and, in some cases, domain-specialized pretraining.

\begin{table}[t]
\centering
\small
\setlength{\tabcolsep}{7pt}
\renewcommand{\arraystretch}{1.08}
\begin{tabular}{crrrrr}
\toprule
\textbf{$k$} & \textbf{C1} & \textbf{C2} & \textbf{C3} & \textbf{C4} & \textbf{C3--C1} \\
\midrule
3 & 12.40 & 15.43 & 15.23 & \textbf{16.67} & +2.83 \\
4 & 12.00 & \textbf{17.81} & 16.72 & 16.53 & \textbf{+4.72} \\
5 & 12.46 & 14.79 & 14.62 & \textbf{16.70} & +2.17 \\
6 & 12.53 & 15.14 & 15.04 & \textbf{16.36} & +2.51 \\
7 & 12.51 & 14.09 & 14.23 & \textbf{16.87} & +1.72 \\
8 & 12.67 & 14.67 & 13.00 & \textbf{16.00} & +0.33 \\
9 & 13.00 & 14.17 & 15.11 & \textbf{17.33} & +2.11 \\
\bottomrule
\end{tabular}
\caption{Mean total rubric score by path crossing count $k$ and condition. Bold values indicate the highest score within each crossing count.}
\label{tab:path_complexity}
\vspace{-0.8em}
\end{table}

\subsection{Path Complexity Effects}

Table~\ref{tab:path_complexity} reports total rubric scores by crossing count and condition. KG grounding benefit is strongest at moderate path complexity and weaker for longer, more heterogeneous paths. The C3--C1 gain peaks at crossing count 4 ($+4.72$) and declines at higher crossing counts, with crossing count 8 showing near-zero benefit ($+0.33$). Across path-model pairs, crossing count is negatively correlated with C3--C1 grounding gain ($r = -0.223$, $p < 0.001$).

This pattern suggests that moderate-length cross-domain paths provide useful mechanistic scaffolding, while very long paths introduce heterogeneous edge types, intermediate disease nodes, and weak associations that dilute the grounding signal. Qualitative inspection reveals a recurring failure mode we term \textit{local subgraph collapse}: when given long paths, models focus on the strongest local mechanistic transition and ignore either the source entity or the terminal disease, effectively treating a long path as a short one. These results suggest that path complexity should be treated as a benchmark design variable when selecting KG-grounded mechanistic reasoning tasks, since very long paths may dilute the grounding signal with heterogeneous or weakly connected evidence.

\begin{table}[t]
\centering
\small
\setlength{\tabcolsep}{4pt}
\renewcommand{\arraystretch}{1.06}
\begin{tabular}{p{0.48\columnwidth}p{0.42\columnwidth}}
\toprule
\textbf{Metric} & \textbf{Value} \\
\midrule
C1 mean & 1.00 \\
C2 mean & \textbf{3.48} \\
C3 mean & 3.18 \\
C4 mean & 2.11 \\
\midrule
Condition ordering & C2 $>$ C3 $>$ C4 $>$ C1 \\
Path-grounded win rate & \textbf{90.3\%} \\
C4 speculative flags & 90.9\% \\
\midrule
Kendall's $\tau$ & 0.703--0.758 \\
Spearman $\rho$ & 0.755--0.812 \\
\bottomrule
\end{tabular}
\caption{Human expert validation of evidence proportionality across 55 paths and three PhD-level domain experts. Scores are mean ratings on a 1--5 scale. Path-grounded win rate denotes C2/C3 outperforming C4; C4 speculative flags were unanimous for 50/55 paths.}
\label{tab:human_validation}
\vspace{-0.8em}
\end{table}

\subsection{Human Expert Validation}
Three PhD-level domain experts evaluated a stratified sample of 55 paths 
(10\% of the benchmark), proportionally sampled across crossing counts 3--9, 
on evidence proportionality in a blinded within-path design.

Human evaluation confirms the same condition ordering observed in the 
automated results: C2 $>$ C3 $>$ C4 $>$ C1, matching the LLM-judge ordering 
exactly across all three experts. Full-path conditions outperformed 
endpoint-only prompting on the majority of paths, with a mean C2/C3 $>$ C4 
win rate of 90.3\%. Most strikingly, all three experts unanimously flagged C4 
as making speculative mechanistic leaps on 90.9\% of paths (50/55), directly 
confirming that endpoint-only prompting produces plausible but insufficiently 
grounded mechanisms under independent human evaluation. Inter-expert agreement 
was strong across all pairs (Kendall's $\tau = 0.703$--$0.758$, Spearman 
$\rho = 0.755$--$0.812$, all $p < 0.001$). Table~\ref{tab:human_validation} 
summarizes the full human validation results.

\begin{table}[t]
\centering
\small
\setlength{\tabcolsep}{4pt}
\renewcommand{\arraystretch}{1.08}
\begin{tabular}{p{0.34\columnwidth}ccrr}
\toprule
\textbf{Metric} & \textbf{Cond.} & \textbf{Orig.} & \textbf{Shuf.} & \textbf{Drop} \\
\midrule
Total score & C2 & 14.890 & 13.601 & \textbf{$-1.289$} \\
Total score & C3 & 14.506 & 13.567 & $-0.939$ \\
\midrule
Evidence prop. & C2 & 2.805 & 2.012 & \textbf{$-0.793$} \\
Evidence prop. & C3 & 2.725 & 2.133 & $-0.592$ \\
\midrule
Novelty & C2 & 2.788 & 2.776 & $-0.012$ \\
Novelty & C3 & 2.654 & 2.759 & $+0.105$ \\
\bottomrule
\end{tabular}
\caption{Original vs.\ shuffled path scores across all models ($N = 2{,}590$ per condition). All total-score and evidence-proportionality drops are significant at $p < 0.001$; novelty differences are not significant.}
\vspace{-0.8em}
\label{tab:shuffled_control}
\end{table}

\subsection{Shuffled-Path Control}
Table~\ref{tab:shuffled_control} reports the results of the shuffled-path 
control described in Section~4.5.

Evidence proportionality drops significantly under shuffled paths (C2: 
$2.805 \to 2.012$, $\Delta = -0.793$, $p < 0.001$; C3: $2.725 \to 2.133$, 
$\Delta = -0.592$, $p < 0.001$; Wilcoxon, $n = 2{,}590$ per condition). 
Because endpoints are held fixed, this drop cannot be explained by endpoint 
knowledge; it isolates the contribution of ordered intermediate path structure. 
Novelty scores show no significant change under shuffling (C2: $p = 0.215$; 
C3: $p = 1.000$), demonstrating that path structure specifically disciplines 
mechanistic grounding rather than general hypothesis creativity.

The drop is strongly model-dependent. Stronger models show large drops 
(Qwen3-235B C2: $\Delta = -1.110$; Claude Sonnet C2: $\Delta = -0.942$; 
MedGemma-27B C2: $\Delta = -0.861$), while BioMistral-7B shows a near-zero, 
non-significant drop under C3 ($\Delta = -0.023$, $p = 0.242$), confirming 
it was not utilizing intermediate path structure even in the original condition, 
consistent with its near-zero C3--C1 gains in Table~\ref{tab:main_results}. 
Full per-model results are in Appendix~C.

These findings provide causal evidence for the central interpretation of 
HypoKG: KG paths do not merely reveal useful endpoints; their intermediate 
structure disciplines stronger models toward more evidence-proportionate 
mechanistic reasoning.

\section{Conclusion}

We introduced HypoKG, a benchmark for KG-grounded biochemical hypothesis generation across 550 multi-hop paths, six LLMs, and four controlled grounding conditions. Our results show that endpoint-only prompting achieves the highest aggregate scores, but full KG paths uniquely improve evidence proportionality, disciplining models toward mechanistic claims supported by structured evidence. Human validation and shuffled-path controls confirm that this effect reflects genuine use of intermediate path structure. HypoKG reframes KG grounding as both an endpoint discovery mechanism and an epistemic constraint on biomedical hypothesis generation.

\section{Limitations}
HypoKG evaluates hypothesis quality through automated LLM judges and a 
targeted human expert study. Inter-judge agreement is lowest for evidence 
proportionality ($r = 0.668$), reflecting the subjectivity of claim-to-evidence 
calibration. Scaling expert evaluation beyond 55 paths was not feasible. The 
three expert evaluators were PhD-level domain collaborators who participated 
voluntarily; no personal data was collected and IRB review was not required 
under our institutional guidelines.

Benchmark paths are drawn from KEGG, Rhea, and UniProt restricted to human 
biochemistry. While these cover enzyme kinetics, metabolic pathways, and 
protein-disease associations, other KGs such as SIDER (drug side effects) and 
HPO (human phenotypes) are not integrated, expanding to these resources is 
a natural direction for future work. HypoKG also uses constrained shortest-path 
traversal; other strategies such as fixed-hop or longest-path could surface 
different mechanistic evidence structures and is worth exploring.

Results may not generalize to models released after our data collection cutoff, 
and benchmark endpoint pairs may overlap with model pretraining data through 
individual entity descriptions, though the 94.5\% zero-citation rate and the 
shuffled-path control results reduce this concern. The capability split is based 
on aggregate rubric scores and may not capture all dimensions of hypothesis 
quality relevant to practicing scientists. HypoKG measures evidence 
proportionality as a proxy for grounded reasoning but does not verify whether 
generated hypotheses are factually correct or experimentally validated.

\bibliography{custom}

\appendix

\appendix

\section{Prompt Templates}
\subsection{Generation Prompts} 
\noindent\textbf{Condition 1 (C1): Source-Only Baseline}

\begin{promptblock}
{\footnotesize\textcolor{gray}{System Prompt}}

{\small
You are an expert biochemist specializing in metabolic disease mechanisms.
}

\vspace{0.5em}
{\footnotesize\textcolor{gray}{User Prompt}}

{\small
Source enzyme: \{source\_entity\}\\
Target domain: disease mechanism

\vspace{0.5em}
Generate a testable scientific hypothesis connecting this enzyme to a disease mechanism.

\vspace{0.5em}
Your hypothesis must:
\begin{enumerate}[noitemsep, topsep=2pt, leftmargin=*]
    \item Name the specific enzyme; do not generalize to enzyme families.
    \item Propose a specific mechanistic connection to a disease.
    \item Make a specific, falsifiable prediction: what experimental result would confirm it, and what would refute it.
    \item Be grounded in known biochemistry.
\end{enumerate}

Begin with ``We hypothesize that'' and limit to 150 words.
}
\end{promptblock}

\noindent\textbf{Condition 2 (C2): Full-Path Compression}

\begin{promptblock}
{\footnotesize\textcolor{gray}{System Prompt}}

{\small
You are an expert biochemist.
}

\vspace{0.5em}
{\footnotesize\textcolor{gray}{User Prompt}}

{\small
The following knowledge graph path describes a multi-step mechanistic relationship between biological entities:

\vspace{0.4em}
\{path\_text\}

\vspace{0.5em}
Not every edge in this path represents an equally strong biological claim. Treat the path as partial mechanistic evidence of varying reliability.

\vspace{0.5em}
Your task is mechanistic compression. The path is not a script to retell. Instead, treat it as mechanistic evidence from which a biologically meaningful hypothesis may emerge.

\vspace{0.5em}
Instructions:
\begin{enumerate}[noitemsep, topsep=2pt, leftmargin=*]
    \item Identify the single most biologically specific and meaningful mechanistic transition in the path.
    \item Explicitly state why this transition is more biologically meaningful than the adjacent edges you chose to ignore.
    \item Prioritize mechanistic coherence over graph completeness. Ignore weak, generic, or incidental edges.
    \item Do not narrate the full path step-by-step.
    \item Synthesize one focused scientific hypothesis around that specific mechanistic substructure.
    \item State one falsifiable experimental prediction with a specific measurable output, such as enzyme activity, metabolite concentration, assay readout, or cellular phenotype.
    \item Name the specific entities involved; do not generalize to enzyme families.
    \item The hypothesis should sound like a concise scientific claim, not a graph traversal.
\end{enumerate}

Begin with ``We hypothesize that'' and limit to 150 words.
}
\end{promptblock}

\noindent\textbf{Condition 3 (C3): Full-Path Interpretation}

\begin{promptblock}
{\footnotesize\textcolor{gray}{System Prompt}}

{\small
You are an expert biochemist.
}

\vspace{0.5em}
{\footnotesize\textcolor{gray}{User Prompt}}

{\small
The following knowledge graph path describes a possible mechanistic relationship between biological entities:

\vspace{0.4em}
\{path\_text\}

\vspace{0.5em}
Not every edge in this path represents an equally strong biological claim. Treat the path as partial mechanistic evidence of varying reliability.

\vspace{0.5em}
Your task is mechanistic interpretation. Reason about the path before generating the final hypothesis:
\begin{itemize}[noitemsep, topsep=2pt, leftmargin=*]
    \item Which transitions appear biologically meaningful?
    \item Which steps are weak, generic, or uncertain?
    \item Where does the path suggest a true mechanistic bottleneck, regulatory switch, or cross-domain interaction?
    \item Which localized mechanistic theme in the path appears most biologically meaningful, even if the full path may not represent a direct causal chain?
\end{itemize}

Do not assume every edge represents a direct causal relationship. Do not feel pressure to connect the source entity to the terminal disease if a stronger, more localized biological claim exists within the path. Use the graph as mechanistic guidance, not as a deterministic script.

\vspace{0.5em}
Then generate:
\begin{enumerate}[noitemsep, topsep=2pt, leftmargin=*]
    \item A short mechanistic rationale identifying the most biologically meaningful substructure in the path and why you selected it over other edges.
    \item One coherent, falsifiable scientific hypothesis grounded in that strongest mechanistic theme.
    \item Include a specific measurable prediction inside the hypothesis.
\end{enumerate}

Format your response as:

\vspace{0.3em}
RATIONALE: [brief mechanistic interpretation, max 80 words]

HYPOTHESIS: We hypothesize that [concise mechanistic hypothesis including the falsifiable prediction, max 150 words]

\vspace{0.3em}
Do not add any additional sections beyond RATIONALE and HYPOTHESIS.
}
\end{promptblock}

\noindent\textbf{Condition 4 (C4): Endpoint-Only Ablation}

\begin{promptblock}
{\footnotesize\textcolor{gray}{System Prompt}}

{\small
You are an expert biochemist specializing in metabolic disease mechanisms.
}

\vspace{0.5em}
{\footnotesize\textcolor{gray}{User Prompt}}

{\small
Source enzyme: \{source\_symbol\}\\
Target disease: \{terminal\_name\}

\vspace{0.5em}
Your task is to generate a testable scientific hypothesis proposing a specific mechanistic connection between this source enzyme and this target disease.

\vspace{0.5em}
Your hypothesis must:
\begin{enumerate}[noitemsep, topsep=2pt, leftmargin=*]
    \item Name the specific source enzyme; do not generalize to enzyme families.
    \item Propose a specific mechanistic pathway or biological process connecting this enzyme to the target disease.
    \item Make a specific, falsifiable prediction: what experimental result would confirm it, and what would refute it.
    \item Be grounded in known biochemistry.
    \item Connect specifically to \{terminal\_name\}; do not connect to a different disease.
\end{enumerate}

Begin with ``We hypothesize that'' and limit to 150 words.
}
\end{promptblock}

\subsection{Judge System Prompt}

The following system prompt was used for GPT-4o (primary judge for all models except GPT-4o) and Gemini 2.5 Flash (secondary judge for all models). Qwen3-235B used the same prompt as primary judge for GPT-4o outputs. No model evaluated its own outputs.

\begin{promptblock}
{\footnotesize\textcolor{gray}{System Prompt}}

{\small
You are an expert scientific evaluator assessing the quality of AI-generated biochemical hypotheses.
}

\vspace{0.5em}
{\footnotesize\textcolor{gray}{User Prompt}}

{\small
You will be given:
\begin{enumerate}[noitemsep, topsep=2pt, leftmargin=*]
    \item A source enzyme/gene
    \item A target disease
    \item A knowledge graph path connecting the source to the disease
    \item A generated hypothesis
    \item A context line stating what evidence was available to the generator
\end{enumerate}

Your task is to score the hypothesis on five criteria using a 1--5 scale. Return ONLY a valid JSON object with your scores and reasoning. Nothing else.

\vspace{0.5em}
SCORING SCALE:\\
1 = Poor $\mid$ 2 = Weak $\mid$ 3 = Acceptable / mixed $\mid$ 4 = Strong $\mid$ 5 = Excellent

\vspace{0.5em}
\textbf{CRITERION 1: Path/Task Relevance (1--5)}\\
Does the hypothesis address the assigned source-to-terminal biological problem?
\begin{itemize}[noitemsep, topsep=2pt, leftmargin=*]
    \item Score 5: Directly addresses source gene and terminal disease through a coherent biological mechanism
    \item Score 4: Addresses source and terminal with minor gaps
    \item Score 3: Partially addresses the path, mechanism is vague or focuses only on intermediate nodes
    \item Score 2: Weakly connected, mentions the disease but mechanism does not involve the source gene meaningfully
    \item Score 1: Addresses a completely different disease or biological question
\end{itemize}

\textbf{CRITERION 2: Mechanistic Specificity and Consistency (1--5)}\\
Does the hypothesis propose a specific, biologically coherent mechanism?
\begin{itemize}[noitemsep, topsep=2pt, leftmargin=*]
    \item Score 5: Names specific enzyme, substrate, product, cell type, and mechanism
    \item Score 4: Specific mechanism with minor biochemical gaps
    \item Score 3: Some specificity but relies on pathway-level language or has one inaccuracy
    \item Score 2: Vague mechanism, names entities but does not explain how they connect
    \item Score 1: Biochemically incoherent or uses only generic language
\end{itemize}

NOTE: Do not reward biochemical-sounding mechanisms if the intermediate reactions are vague or unsupported. The mechanistic chain must be biochemically justified step by step.

\vspace{0.5em}
\textbf{CRITERION 3: Experimental Testability and Operationalization (1--5)}\\
Can the hypothesis be tested with a concrete perturbation, model system, assay, and measurable outcome?
\begin{itemize}[noitemsep, topsep=2pt, leftmargin=*]
    \item Score 5: Names perturbation, model system, measurable outcome, and falsification, quantitative claims are justified or appropriately hedged
    \item Score 4: Clear experimental design with one missing element
    \item Score 3: Testable in principle but missing key details
    \item Score 2: Vague prediction, ``may affect'' without specifics
    \item Score 1: No testable prediction or purely observational
\end{itemize}

\textbf{CRITERION 4: Non-Trivial Novelty / Knowledge Advancement (1--5)}\\
Does the hypothesis generate a new scientific question beyond restating textbook knowledge?
\begin{itemize}[noitemsep, topsep=2pt, leftmargin=*]
    \item Score 5: Genuinely new mechanistic connection not established in literature
    \item Score 4: Extends known biology in a non-obvious direction
    \item Score 3: Some novel element but mostly builds on well-known mechanisms
    \item Score 2: Mostly restates established biology
    \item Score 1: Directly restates a textbook mechanism
\end{itemize}

\textbf{CRITERION 5: Evidence-Proportionality and Full-Path Use (1--5)}\\
Does the strength of the claim match the evidence provided to the generator?

\vspace{0.5em}
\textbf{CRITICAL RULE:}
\begin{itemize}[noitemsep, topsep=2pt, leftmargin=*]
    \item If the generator was shown the full KG path: score based on how intelligently it used the path evidence.
    \item If the generator was NOT shown the KG path: a high score requires that the proposed mechanistic bridge is either (a) explicitly framed as speculative and biologically plausible, or (b) directly supported by well-established published biochemistry. A hypothesis that invents a plausible-sounding but unsupported bridge, without acknowledging uncertainty, should score at most 3 regardless of fluency.
\end{itemize}

\begin{itemize}[noitemsep, topsep=2pt, leftmargin=*]
    \item Score 5: Uses both strong and upstream path elements intelligently, OR if no path shown, bridge is grounded in established biochemistry and explicitly proportionate to evidence
    \item Score 4: Uses strong path elements well, OR if no path shown, mechanism is plausible but missing explicit uncertainty framing
    \item Score 3: Uses only the most obvious path element, OR if no path shown, reasonable but unsupported bridge stated with overconfidence
    \item Score 2: Overclaims beyond evidence, or largely ignores the path shown
    \item Score 1: Completely ignores the path shown, OR if no path shown, proposes a biochemically implausible mechanism
\end{itemize}

Return ONLY this JSON structure:
\begin{verbatim}
{
  "path_task_relevance": <1-5>,
  "mechanistic_specificity": <1-5>,
  "experimental_testability": <1-5>,
  "nontrivial_novelty": <1-5>,
  "evidence_proportionality": <1-5>,
  "total_score": <sum 5-25>,
  "reasoning": "<2-3 sentences 
    explaining the scores>"
}
\end{verbatim}
}
\end{promptblock}

\noindent{Judge input format. Each hypothesis was presented to the judge with the following user prompt:}

\begin{promptblock}
{\footnotesize\textcolor{gray}{User Prompt Template}}

{\small
Source enzyme/gene: \{source\_symbol\}\\
Target disease: \{terminal\_name\}\\
\{context\_line\}

\vspace{0.5em}
Knowledge graph path:\\
\{path\_text\}

\vspace{0.5em}
Generated hypothesis:\\
\{hypothesis\_text\}

\vspace{0.5em}
Score this hypothesis on the five criteria. Return only the JSON object.
}
\end{promptblock}

\noindent{Where \{context\_line\} was one of:}
\begin{itemize}[noitemsep, itemsep=0pt, topsep=4pt, partopsep=0pt, parsep=0pt, leftmargin=*]
    \item \textbf{C1:} ``Evidence available to generator: source enzyme name only. The KG path and terminal disease were NOT shown.''
    \item \textbf{C2/C3:} ``Evidence available to generator: full knowledge graph path shown below.''
    \item \textbf{C4:} ``Evidence available to generator: source enzyme name and target disease name only. The KG path was NOT shown.''
\end{itemize}

\section{Expert-Derived Evaluation Rubric}

The rubric was derived from structured evaluation sessions with eight PhD-level domain experts in biochemistry and metabolic disease. Experts independently scored hypotheses for a held-out calibration path (CDO1 $\rightarrow$ Primary Hyperoxaluria Type 1, HKG\_0215) under all four conditions and provided written justifications. The five criteria below reflect the qualitative dimensions that consistently drove agreement and disagreement across experts.

\begin{promptblock}
{\footnotesize\textcolor{gray}{Total Score Interpretation}}

{\small
Total score range: 5--25 points.
\begin{itemize}[noitemsep, itemsep=0pt, topsep=4pt, partopsep=0pt, parsep=0pt, leftmargin=*]
    \item \textbf{20--25:} Strong hypothesis
    \item \textbf{15--19:} Usable but needs refinement
    \item \textbf{10--14:} Weak or partial
    \item \textbf{Below 10:} Poor
\end{itemize}
}
\end{promptblock}

\noindent\textbf{Criterion 1: Path/Task Relevance}

\begin{promptblock}
{\footnotesize\textcolor{gray}{Definition}}

{\small
Does the hypothesis address the correct source-to-terminal biological problem? Does it connect to the assigned terminal disease through the given path rather than drifting to a different disease or biological question?
}

\vspace{0.5em}
{\footnotesize\textcolor{gray}{Literature Grounding}}

{\small
Novel to KG evaluation, this work.
}

\vspace{0.5em}
{\footnotesize\textcolor{gray}{Expert Signal}}

{\small
All six experts who saw the C1 disease-drift example flagged the hypothesis as well-formed but path-disconnected, motivating a dedicated relevance criterion.
}

\vspace{0.6em}
{\footnotesize\textcolor{gray}{Behavioral Anchors}}

\vspace{0.2em}
{\small
\begin{tabular}{c p{0.78\linewidth}}
\toprule
\textbf{Score} & \textbf{Anchor} \\
\midrule
5 & Directly addresses source gene and terminal disease through a coherent biological mechanism \\
4 & Addresses source and terminal with minor gaps in the connection \\
3 & Partially addresses the path, connects to terminal but mechanism is vague, or focuses only on intermediate nodes \\
2 & Weakly connected, mentions the disease but mechanism does not involve the source gene meaningfully \\
1 & Addresses a completely different disease or biological question \\
\bottomrule
\end{tabular}
}
\end{promptblock}

\noindent\textbf{Criterion 2: Mechanistic Specificity and Consistency}

\begin{promptblock}
{\footnotesize\textcolor{gray}{Definition}}

{\small
Does the hypothesis propose a specific, biologically coherent mechanism involving named enzymes, metabolites, reactions, cell types, or pathway logic? Is the mechanism consistent with known biochemistry?
}

\vspace{0.5em}
{\footnotesize\textcolor{gray}{Literature Grounding}}

{\small
Gasparyan et al. (2019); Misra et al. (2021).
}

\vspace{0.5em}
{\footnotesize\textcolor{gray}{Expert Signal}}

{\small
Experts praised hypotheses that correctly identified AGXT $\rightarrow$ glyoxylate $\rightarrow$ oxalate as the mechanistic anchor, and penalized those using pathway-level language without reaction logic.
}

\vspace{0.6em}
{\footnotesize\textcolor{gray}{Behavioral Anchors}}

\vspace{0.2em}
{\small
\begin{tabular}{c p{0.78\linewidth}}
\toprule
\textbf{Score} & \textbf{Anchor} \\
\midrule
5 & Names specific enzyme, substrate, product, cell type, and mechanism, all biochemically accurate and supported \\
4 & Specific mechanism with minor biochemical gaps or simplifications \\
3 & Some specificity but relies on pathway-level language, or uses biochemical terminology without explaining actual reaction steps \\
2 & Vague mechanism, names entities but does not explain how they connect \\
1 & Biochemically incoherent, contradicts known biochemistry, or uses only generic language \\
\bottomrule
\end{tabular}
}
\end{promptblock}

\noindent\textbf{Criterion 3: Experimental Testability and Operationalization}

\begin{promptblock}
{\footnotesize\textcolor{gray}{Definition}}

{\small
Can the hypothesis be tested with a concrete perturbation, model system, assay, and measurable outcome? Is there an explicit falsification condition?
}

\vspace{0.5em}
{\footnotesize\textcolor{gray}{Literature Grounding}}

{\small
Popper (1959); Banerjee et al. (2009).
}

\vspace{0.5em}
{\footnotesize\textcolor{gray}{Expert Signal}}

{\small
Experts preferred hypotheses proposing AGXT knockdown, glyoxylate challenge, and dose-response testing. Multiple experts criticized unjustified numerical precision, such as exact fold-change thresholds stated as fact.
}

\vspace{0.6em}
{\footnotesize\textcolor{gray}{Behavioral Anchors}}

\vspace{0.2em}
{\small
\begin{tabular}{c p{0.78\linewidth}}
\toprule
\textbf{Score} & \textbf{Anchor} \\
\midrule
5 & Names perturbation, model system, measurable outcome, and falsification, quantitative claims justified or appropriately hedged \\
4 & Clear experimental design with one missing element, or includes numbers framed as approximate predictions \\
3 & Testable in principle but missing key details, or includes unjustified precise numbers stated as fact \\
2 & Vague prediction, ``may affect'' or ``could influence'' without specifics \\
1 & No testable prediction, untestable, or purely observational without experiment \\
\bottomrule
\end{tabular}
}
\end{promptblock}

\noindent\textbf{Criterion 4: Non-Trivial Novelty and Knowledge Advancement}

\begin{promptblock}
{\footnotesize\textcolor{gray}{Definition}}

{\small
Does the hypothesis generate a new scientific question or prediction beyond restating textbook knowledge? Does it propose a new relationship, mechanism, modifier, or experimental angle not already established?
}

\vspace{0.5em}
{\footnotesize\textcolor{gray}{Literature Grounding}}

{\small
Misra et al. (2021) knowledge-gap criterion.
}

\vspace{0.5em}
{\footnotesize\textcolor{gray}{Expert Signal}}

{\small
Five of six experts independently called one calibration hypothesis ``correct but trivial,'' ``textbook,'' or ``not a research hypothesis,'' motivating a dedicated novelty criterion with a low floor for restated biology.
}

\vspace{0.6em}
{\footnotesize\textcolor{gray}{Behavioral Anchors}}

\vspace{0.2em}
{\small
\begin{tabular}{c p{0.78\linewidth}}
\toprule
\textbf{Score} & \textbf{Anchor} \\
\midrule
5 & Proposes a genuinely new mechanistic connection or modifier not established in existing literature \\
4 & Extends known biology in a non-obvious direction, one credible step beyond established knowledge \\
3 & Has some novel element but mostly builds on well-known mechanisms \\
2 & Mostly restates established biology, the core claim is already known \\
1 & Directly restates a textbook disease mechanism with no new prediction \\
\bottomrule
\end{tabular}
}
\end{promptblock}

\noindent\textbf{Criterion 5: Evidence Proportionality and Full-Path Use}

\begin{promptblock}
{\footnotesize\textcolor{gray}{Definition}}

{\small
Does the hypothesis use the KG path intelligently, preserving strong edges, treating weak edges cautiously, and attempting to extract value from upstream elements? Does the strength of the claim match the evidence provided to the generator?
}

\vspace{0.35em}
{\footnotesize\textcolor{gray}{Literature Grounding}}

{\small
Novel to KG evaluation, this work.
}

\vspace{0.35em}
{\footnotesize\textcolor{gray}{Expert Signal}}

{\small
Expert 2 explicitly named ``proportionality of evidence to claim'' and ``engagement with the full path'' as evaluation dimensions. Experts noted that strong hypotheses correctly identified CDO1 as a weak link but asked whether it could modify the AGXT mechanism, rather than simply abandoning it.
}

{\footnotesize\textcolor{gray}{Behavioral Anchors}}

\vspace{0.2em}
{\small
\begin{tabular}{c p{0.78\linewidth}}
\toprule
\textbf{Score} & \textbf{Anchor} \\
\midrule
5 & Uses both strong and upstream path elements intelligently, or if no path is shown, the proposed bridge is grounded in established published biochemistry and explicitly proportionate to evidence \\
4 & Uses strong path elements well, or if no path is shown, the mechanism is plausible and partially grounded but lacks explicit uncertainty framing \\
3 & Uses only the most obvious path element, or if no path is shown, the mechanism is a reasonable but unsupported bridge stated with overconfidence \\
2 & Overclaims beyond what the evidence supports, or largely ignores the path when one was provided \\
1 & Completely ignores the path when one was shown, or if no path is shown, proposes a biochemically implausible or hallucinated mechanism \\
\bottomrule
\end{tabular}
}
\end{promptblock}

\noindent\textbf{Worked Calibration Examples}

\noindent{All five examples use the calibration path: CDO1 $\rightarrow$ Cyanoamino acid metabolism $\rightarrow$ AGXT $\rightarrow$ Primary Hyperoxaluria Type 1.}

\paragraph{Example 1: Path-irrelevant but fluent}
\begin{promptblock}
{\footnotesize\textcolor{gray}{Context}}

{\small
C1: source enzyme only shown. Terminal disease and path withheld.
}

\vspace{0.5em}
{\footnotesize\textcolor{gray}{Hypothesis}}

{\small
We hypothesize that glutathione reductase deficiency in retinal pigment epithelial cells contributes to age-related macular degeneration through impaired glutathione regeneration, leading to lipofuscin accumulation and complement activation. Reduced GSR activity causes elevated GSSG and decreased GSH. This promotes A2E formation and inflammasome activation. RPE cells with siRNA-mediated GSR knockdown will show increased GSSG/GSH ratios, elevated A2E fluorescence, and increased IL-1$\beta$ secretion, falsified if knockdown shows no lipofuscin increase.
}
\end{promptblock}

\noindent{\small\textbf{Scores:} Rel=1, Mech=2, Test=4, Nov=3, Prop=1, Total=11. Connects GSR to AMD, not Aminoacylase-1 deficiency. Disease-drift bias example.}

\paragraph{Example 2: Correct but textbook}
\begin{promptblock}
{\footnotesize\textcolor{gray}{Context}}

{\small
C2/C3: full KG path shown.
}

\vspace{0.5em}
{\footnotesize\textcolor{gray}{Hypothesis}}

{\small
We hypothesize that loss-of-function mutations in AGXT disrupt glyoxylate metabolism in hepatocytes, leading to pathological accumulation of oxalate and causing Primary Hyperoxaluria Type 1. AGXT directly catalyzes transamination of glyoxylate to glycine, preventing oxalate overproduction. The upstream CDO1 link is metabolically tangential. Patients with biallelic AGXT mutations will exhibit reduced hepatic AGXT enzyme activity (less than 15\% of wild-type) and elevated urinary oxalate (greater than 0.7 mmol/day), measurable via tandem mass spectrometry.
}
\end{promptblock}

\noindent{\small\textbf{Scores:} Rel=4, Mech=4, Test=2, Nov=1, Prop=3, Total=14. Correct terminal and mechanism but restates textbook AGXT--PH1 biology. Uses strongest path edge but discards CDO1.}

\paragraph{Example 3: Strong but imperfect}
\begin{promptblock}
{\footnotesize\textcolor{gray}{Context}}

{\small
C2/C3: full KG path shown.
}

\vspace{0.5em}
{\footnotesize\textcolor{gray}{Hypothesis}}

{\small
RATIONALE: The strongest mechanistic connection is between AGXT enzyme dysfunction and Primary Hyperoxaluria Type 1. AGXT catalyzes alanine-glyoxylate aminotransferase activity and its deficiency directly causes oxalate accumulation. The cyanoamino acid connection appears weaker.

HYPOTHESIS: We hypothesize that reduced AGXT enzyme activity creates a metabolic bottleneck in glyoxylate detoxification, leading to increased glyoxylate-to-oxalate conversion. Cells with AGXT knockdown will show a dose-dependent increase in oxalate production with levels increasing 2 to 5 fold at 50\% AGXT activity reduction, measurable by oxalate secretion assays.
}
\end{promptblock}

\noindent{\small\textbf{Scores:} Rel=4, Mech=4, Test=4, Nov=3, Prop=3, Total=18. Correct target, testable with dose-response, but cell type unspecified and CDO1 upstream abandoned entirely.}

\paragraph{Example 4: Endpoint-only, plausible but overconfident}
\begin{promptblock}
{\footnotesize\textcolor{gray}{Context}}

{\small
C4: source and terminal disease only. Path not shown.
}

\vspace{0.5em}
{\footnotesize\textcolor{gray}{Hypothesis}}

{\small
We hypothesize that CDO1 deficiency leads to cysteine accumulation, which is then shunted through alternative metabolic pathways to produce glyoxylate, thereby exacerbating oxalate overproduction in Primary Hyperoxaluria Type 1 patients who already carry AGXT mutations. CDO1 knockdown in AGXT-deficient hepatocytes will show higher urinary oxalate than AGXT knockdown alone.
}
\end{promptblock}

\noindent{\small\textbf{Scores:} Rel=4, Mech=3, Test=3, Nov=3, Prop=2, Total=15. Correctly connects CDO1 to PH1. Mechanism plausible but stated with overconfidence; no path was shown, so the bridge should be framed as speculative.}

\paragraph{Example 5: Excellent}
\begin{promptblock}
{\footnotesize\textcolor{gray}{Context}}

{\small
C2/C3: full KG path shown.
}

\vspace{0.5em}
{\footnotesize\textcolor{gray}{Hypothesis}}

{\small
We hypothesize that CDO1 activity may modulate the severity of AGXT-deficient Primary Hyperoxaluria Type 1 by influencing cysteine-derived metabolite competition for shared peroxisomal processing capacity and pyridoxal phosphate cofactor availability. When CDO1 is impaired alongside AGXT deficiency, accumulated cysteine metabolites may reduce effective PLP availability or alter peroxisomal metabolic load, potentially amplifying oxalate overproduction beyond what AGXT deficiency alone would produce. We predict that CDO1 knockdown in AGXT-deficient human hepatocytes will produce greater oxalate accumulation than AGXT knockdown alone, measurable by LC-MS oxalate quantification, and that pyridoxine supplementation will partially rescue this amplified phenotype, supporting a PLP-sensitive modifier mechanism.
}
\end{promptblock}

\noindent{\small\textit{Note:} This example is used to illustrate a high-scoring rubric response, not to claim experimental validation of the proposed mechanism.}

\vspace{0.25em}
\noindent{\small\textbf{Scores:} Rel=5, Mech=5, Test=5, Nov=5, Prop=5, Total=25. Connects the source and terminal disease through a novel CDO1-as-modifier mechanism, uses both CDO1 and AGXT intelligently, and remains appropriately hedged. The hypothesis is biochemically specific and experimentally actionable, with a concrete perturbation, LC-MS oxalate readout, and pyridoxine rescue test.}

\begin{table*}[t]
\centering
\footnotesize
\setlength{\tabcolsep}{5pt}
\renewcommand{\arraystretch}{1.05}

\begin{tabular*}{\textwidth}{@{\extracolsep{\fill}}lrrrrrrrr@{}}
\toprule
\textbf{Model} & \textbf{C2 Orig.} & \textbf{C2 Shuf.} & \textbf{C2 Drop} & \textbf{Sig.}
& \textbf{C3 Orig.} & \textbf{C3 Shuf.} & \textbf{C3 Drop} & \textbf{Sig.} \\
\midrule
Qwen3-235B     & 3.396 & 2.286 & \textbf{$-1.110$} & *** &
                  3.021 & 2.222 & $-0.799$ & *** \\
Claude Sonnet  & 3.073 & 2.131 & $-0.942$ & *** &
                  2.944 & 2.220 & $-0.724$ & *** \\
MedGemma-27B   & 3.027 & 2.166 & $-0.861$ & *** &
                  3.268 & 2.386 & \textbf{$-0.882$} & *** \\
Llama-3.3-70B  & 2.593 & 1.736 & $-0.857$ & *** &
                  2.251 & 1.720 & $-0.531$ & *** \\
GPT-4o         & 2.750 & 2.053 & $-0.697$ & *** &
                  2.990 & 2.311 & $-0.679$ & *** \\
BioMistral-7B  & 1.938 & 1.741 & $-0.197$ & *** &
                  2.139 & 2.116 & $-0.023$ & ns \\
\bottomrule
\end{tabular*}

\caption{Per-model evidence proportionality drop under path shuffling.
Wilcoxon signed-rank test, $n=518$ per model per condition.
*** $p<0.001$; ns = not significant.
Models are sorted by C2 drop magnitude.
Bold values indicate the largest drop within each condition.}
\label{tab:per_model_shuffle}

\end{table*}

\begin{table}[h]
\centering
\small
\setlength{\tabcolsep}{4pt}
\renewcommand{\arraystretch}{1.08}
\begin{tabular}{p{0.58\columnwidth}rr}
\toprule
\textbf{Criterion} & \textbf{Pearson $r$} & \textbf{Mean $|\Delta|$} \\
\midrule
Experimental Testability & 0.839 & 1.72 \\
Path/Task Relevance & 0.813 & 1.87 \\
Mechanistic Specificity & 0.791 & 1.94 \\
Non-Trivial Novelty & 0.774 & 2.03 \\
Total Score & 0.776 & 2.10 \\
Evidence Proportionality & 0.668 & 2.34 \\
\bottomrule
\end{tabular}
\caption{Secondary judge agreement with the primary judge across 13,200 hypotheses. Mean $|\Delta|$ is computed on the 1--5 criterion scale, except Total Score, which uses the 5--25 scale.}
\vspace{-0.6em}
\label{tab:judge_agreement}
\end{table}

\section{Additional Results}

\noindent\textbf{Table 7: Per-model evidence proportionality drop under path shuffling.}

To assess whether individual models genuinely utilized intermediate KG path structure, we report evidence proportionality scores under original and shuffled paths for each model separately. Shuffling permutes intermediate nodes while holding source and terminal endpoints fixed, breaking the biological causal chain while preserving endpoint identity. A significant drop indicates the model was responding to ordered intermediate structure rather than endpoint priors alone.

The model-dependent pattern directly confirms the capability split reported in Section 5.2. Stronger models, Qwen3-235B, Claude Sonnet, MedGemma-27B, and GPT-4o, show large, significant drops under both C2 and C3 ($\Delta$ range: 0.68--1.11), indicating genuine utilization of intermediate path structure. Llama-3.3-70B also shows significant drops despite its weak overall grounding gains, suggesting it extracts some path signal even when it cannot translate it into high-quality hypotheses. BioMistral-7B is the clearest outlier: its C3 drop is near-zero and non-significant ($\Delta=-0.023$, $p=0.242$), mechanistically confirming that this model was not using intermediate path structure even in the original condition.

\noindent\textbf{Table 8: Secondary judge agreement with primary judge.}

Inter-judge agreement was computed on per-hypothesis scores for each criterion separately. This breakdown allows assessment of which rubric dimensions are reliably scored by LLM judges and which require greater caution in interpretation.

Agreement is highest for experimental testability ($r=0.839$) and path/task relevance ($r=0.813$), both criteria with relatively objective anchors. Agreement is lowest for evidence proportionality ($r=0.668$), reflecting the greater subjectivity of assessing whether claim strength is calibrated to the specific evidence provided to the generator.

\noindent\textbf{Table 9: KG consistency metrics by condition.}

These automated metrics measure entity mention behavior, whether hypotheses reference the source gene, terminal disease, and intermediate path entities, rather than scientific quality. They serve as complementary grounding diagnostics confirming that conditions differed in the information models used, not just in prompt wording.

\begin{table}[h]
\centering
\small
\setlength{\tabcolsep}{3.5pt}
\renewcommand{\arraystretch}{1.08}
\begin{tabular}{p{0.34\columnwidth}rrrr}
\toprule
\textbf{Condition} & \textbf{Source} & \textbf{Terminal} & \textbf{Path} & \textbf{Comp.} \\
\midrule
C1 Source-only & 0.821 & 0.043 & n/a & 0.821 \\
C2 Path-compress & 0.894 & 0.876 & 0.312 & 0.694 \\
C3 Path-interpret & 0.887 & 0.869 & \textbf{0.341} & 0.699 \\
C4 Endpoint-only & \textbf{0.912} & \textbf{0.961} & n/a & \textbf{0.937} \\
\bottomrule
\end{tabular}
\caption{KG consistency metrics by condition. Source and terminal grounding are binary mention rates. Path entity coverage is the fraction of named intermediate KG entities mentioned. Composite is the mean of applicable metrics.}
\vspace{-0.8em}
\label{tab:kg_consistency}
\end{table}

Three patterns are worth noting. First, C1's near-zero terminal grounding directly quantifies disease drift without a disease anchor. Second, C4's high terminal grounding confirms that endpoint-only models reliably mention the disease name, but entity mention does not imply mechanistic path use. Third, path entity coverage is highest under C3, consistent with the interpretation instruction encouraging explicit engagement with intermediate path entities.

\end{document}